\documentclass[conference]{IEEEtran}
\IEEEoverridecommandlockouts
\usepackage{cite}
\usepackage{amsmath,amssymb,amsfonts}
\usepackage{algorithmic}
\usepackage{graphicx}
\usepackage{textcomp}
\usepackage{xcolor}
\usepackage{graphicx}
\usepackage{amsmath}
\usepackage{amssymb}
\usepackage{booktabs}
\usepackage{float}
\usepackage{wrapfig}
\usepackage{longtable}
\usepackage{algorithm,algorithmic}
\usepackage{hyperref}
\usepackage{longtable} 
\usepackage{array} 
\usepackage[capitalize]{cleveref}
\crefname{section}{Sec.}{Secs.}
\Crefname{section}{Section}{Sections}
\Crefname{table}{Table}{Tables}
\crefname{table}{Tab.}{Tabs.}

\def\BibTeX{{\rm B\kern-.05em{\sc i\kern-.025em b}\kern-.08em
    T\kern-.1667em\lower.7ex\hbox{E}\kern-.125emX}}
\begin{document}

\title{SparrowVQE: Visual Question Explanation for Course Content Understanding
}

\author{\IEEEauthorblockN{Jialu Li, Manish Kumar Thota, Ruslan Gokhman, Radek Holik, and Youshan Zhang}
\IEEEauthorblockA{\textit{Artificial Intelligence} \\
\textit{Graduate Computer Science and Engineering Department}\\
Yeshiva University, NY, USA \\
\{jli10,mthota,rgokhman,rholik\}@mail.yu.edu, youshan.zhang@yu.edu}\\
}

\maketitle

\begin{abstract}
Visual Question Answering (VQA) research seeks to create AI systems to answer natural language questions in images, yet VQA methods often yield overly simplistic and short answers. This paper aims to advance the field by introducing Visual Question Explanation (VQE), which enhances the ability of VQA to provide detailed explanations rather than brief responses and address the need for more complex interaction with visual content. We first created an MLVQE dataset from a 14-week streamed video machine learning course, including 885 slide images, 110,407 words of transcripts, and 9,416 designed question-answer (QA) pairs. Next, we proposed a novel SparrowVQE, a small 3 billion parameters multimodal model. We trained our model with a three-stage training mechanism consisting of multimodal pre-training (slide images and transcripts feature alignment), instruction tuning (tuning the pre-trained model with transcripts and QA pairs), and domain fine-tuning (fine-tuning slide image and QA pairs). Eventually, our SparrowVQE can understand and connect visual information using the SigLIP model with transcripts using the Phi-2 language model with an MLP adapter. Experimental results demonstrate that our SparrowVQE achieves better performance in our developed MLVQE dataset and outperforms state-of-the-art methods in the other five benchmark VQA datasets. The source code is available at \url{https://github.com/YoushanZhang/SparrowVQE}.
\end{abstract}



\section{Introduction}
\label{sec:intro}
Visual Question Answering (VQA) is an interdisciplinary problem that combines computer vision with natural language processing in answering questions regarding images, aiming to recognize and localize objects and information presented in a context. It makes positive differences in various applications, such as aiding visually impaired people, supporting educational tools, and developing user interfaces of human-computer interaction~\cite{barra2021visual}. 

\begin{figure}[h!]
  \centering
  \includegraphics[width=1.1\linewidth]{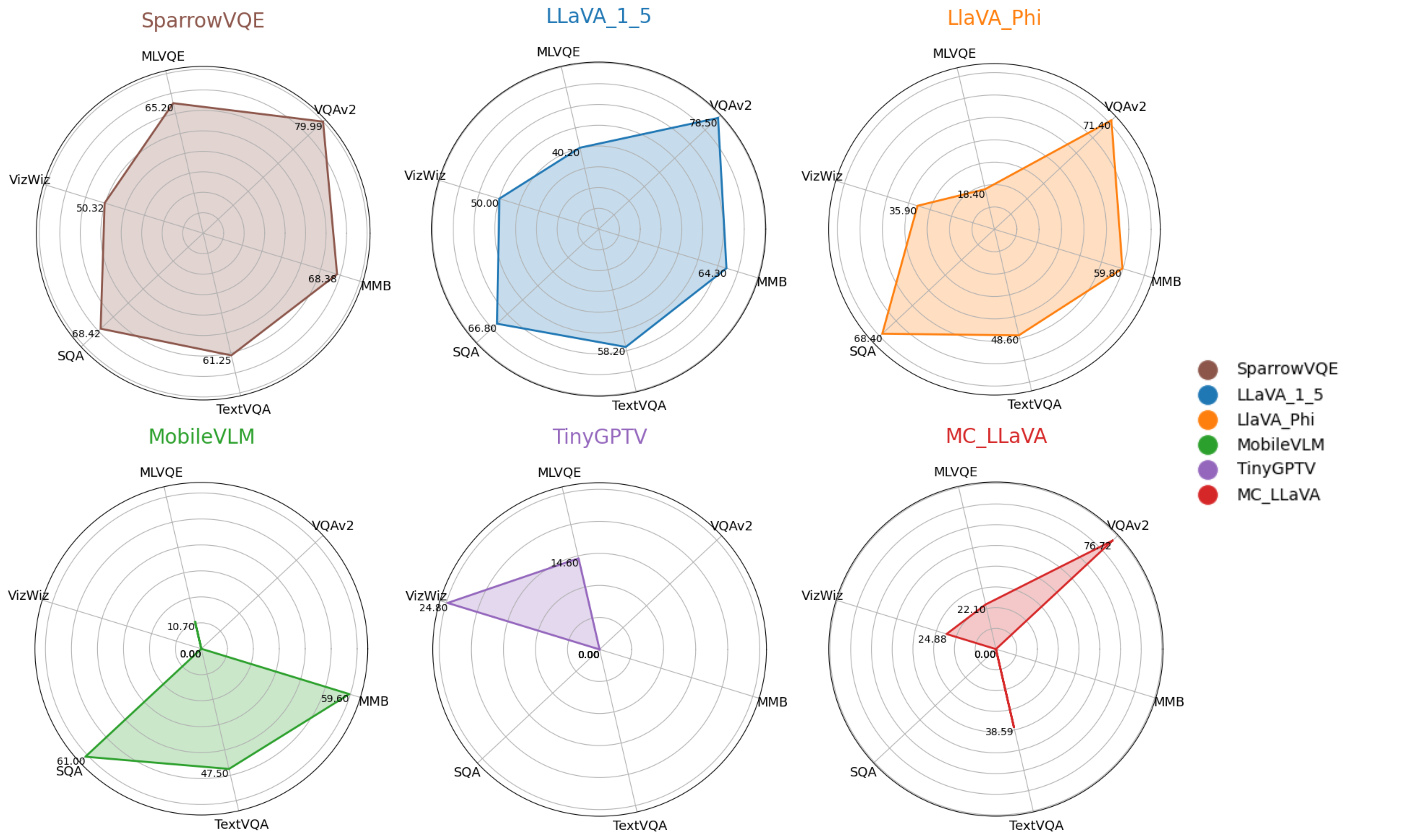}
  \caption{Our SparrowVQE matches the performance of 7B models in numerous visual language tasks, standing out from general-purpose text and visual language models.}
  \vspace{-0.7cm}
  \label{fig:overall}
\end{figure}

A major challenge for the development of VQA is the large diversity of questions that can be asked, from simple identifying tasks, such as ``What is in the picture?", to complex queries requiring sophisticated comprehension and inferential reasoning of the relationships and stories in the visual content. Applying VQA in educational settings, specifically machine learning (ML) lectures, can be more complex, as ML lectures often include complex diagrams, mathematical formulas, and dense textual information.

Traditional educational resources often fail to provide the engagement and context-aware assistance necessary for effective learning. This gap results in difficulties in bridging theoretical knowledge with practical understanding in the educational context. Recently, several education chatbots or VQA systems were developed to improve students' learning experience in different education levels~\cite{gupta2023eduvi,he2017educational,xiong2020ta,sophia2021edubot,lin2023research,cheng2023application}.  However, the models' performance varies with preliminary evaluation and mostly produce short answers instead of detailed explanation of questions. The educational VQAs still face the problem of limited training data and over-simplistic answers produced.

In this paper, we propose an MLVQE dataset for model training in the machine learning setting, specifically to achieve automatic teaching. We also propose a SparrowVQE model to enrich the VQA application in education. It caters to ML learners by allowing them to ask questions about visual content directly. 
Our work directly contributes to the improvement of effectiveness and personalizing of learning experiences. We improve our model's efficiency in interpreting slide-text/transcript pairs, making it outperform state-of-the-art models in VQA tasks in different contents, as shown in Fig.~\ref{fig:overall}. Our model has the potential to be applied in other educational domains that require training with more datasets. 

\section{Related work}
Visual question-answering (VQA) is an emerging AI technique in the recent decade that combines computer vision with natural language processing\cite{zou2020survey}. Initial VQA models were simple in architecture, and limited datasets restricted their performance. 
Malinowski and Fritz~\cite{malinowski2014multi} proposed one of the first open-ended datasets for image question answering with 894 object classes and more than 12,000 QA pairs based on the NYU-Depth V2 dataset\cite{silberman2012indoor}. This dataset later became a foundation for subsequent research. The VQA field evolved following the increasing availability of benchmark datasets. The VQA v1.0\cite{agrawal2016vqa,de2023visual}, with the extraction from Microsoft COCO-VQA dataset \cite{lin2015microsoft} consisted of diverse and complex question-answer pairs that challenged the limits of VQA models, and paid more attention to practical contexts. Antol et al.\cite{agrawal2016vqa} expanded the MS COCO dataset with an additional abstract scene dataset that contained 50,000 scenes\cite{zitnick2013bringing}, contributing to the improved ability of image understanding and complex reasoning. 

With the improvement of VQA systems' comprehension and reasoning capacities, they have been applied in various fields, such as medical\cite{demirhan2023survey} and education\cite{barra2021visual}.  Medical Visual Question Answering (Med-VQA)\cite{WANG2022102346} is a critical frontier in applying artificial intelligence to medical image interpretation through automated question and answering systems. 
The recent VQA dataset in 2017 contains 204,721 images, 1,105,904 questions, and 11,059,040 ground truth answers not specific to any domain, publicly available at VQA v2.0\cite{goyal2017making,demirhan2023survey}. VQA-Med datasets~\cite{ben2021overview} and VQA-RAD Dataset\cite{lau2018dataset} focused on images in multiple radiology domains. Fusion-based methods are commonly used in medical VQA~\cite{depeursinge2010fusion}, including combining features at the system input with intermedia query expansion, internally to the system with early fusion, or at the output of the system with late fusion. Attention-based methods further evolved the performance of medical VQA. Pan et al.\cite{pan2021muvam} introduced MuVAM,  highlighting the inclusion of a multi-view attention mechanism that included Image-to-Question (I2Q) attention and Word-to-Text (W2T) attention, correlating the question with images and words, and a composite loss for more accurate answer predictions on VQA-RAD and VQA-RADPh datasets. The model achieved better effectiveness than state-of-the-art methods. 

Compared to the medical field, the application of VQA in education is relatively new. Datasets in this genre include EgoVQA~\cite{9022046} with cooking video QA pairs and How2QA~\cite{li2020hero} with science video clips and questions. Sophia et al.\cite{sophia2021edubot} presented a student chatbot system that integrated RNNs and CNNs for VQA language and image processing, and Dialogflow for connecting to external agents. The chatbot system is compatible with any online teaching platform. Gupta et al.\cite{gupta2023eduvi} introduced an EDUVI system with VQA and image caption modules, which used CNNs and LSTM models for image and text extraction and caption. The model was trained on images from the 4th standard E.V.S textbook and proved useful in improving primary-level students' learning and thinking capacities. 
Lin~\cite{lin2023research} proposed a multiscale fusion deep learning method and an improved mixed attention mechanism (spatial domain attention and channel attention) for a college student VQA learning system. 

Large language models have also been applied in the VQA task. 
CLIP~\cite{radford2021learning} showed the ability to understand images based on natural language descriptions in VQA systems by providing robust image-text representations. BLIP~\cite{li2023blip} diverged from traditional language models, like a generative pre-trained transformer (GPT), to be a multimodal model tailored for operations that require the analysis of both text and images. LLaVA represented a cost-effective approach to create flexible multimodal assistance~\cite{liu2023llava}, which combined a vision encoder with Vicuna to provide comprehensive insights for both visual and language-based information. Pix2Struct~\cite{lee2023pix2struct} fine-tuned across broad tasks and datasets, including work on image captioning, VQA, diverse sources like books, charts, diagrams, and labeling UI elements. SigLIP\cite{zhai2023sigmoid} proposed substituting the loss function in CLIP with a straightforward pairwise sigmoid loss, which could improve the efficiency of language-image pre-training. However, their application in the education domain is still inadequate.

To address the lack of training data for education VQA systems, especially in AI tutoring,  which contains more sophisticated contexts, inferences, and possibly more interactions, we propose a dataset based on recorded machine learning lectures with open-ended question explanation pairs named MLVQE.  To handle complex questions in course materials, our SparrowVQE model has a three-stage training mechanism: (1). multimodal pre-training (slide image and transcripts feature alignment), (2). instruction tuning (tuning the pre-trained model with transcripts and QA pairs), and (3) domain fine-tuning (fine-tuning slide image and QA pairs). Our 3B parameters SparrowVQE can understand and connect visual information using the SigLIP model with transcripts using the Phi-2 language model with an MLP adapter. It can be deployed to mobile devices as an effective tool for educational content production. The proposal of our model will enable more interactive learning environment for students 

\begin{figure}[H] 
  \centering
  \includegraphics[width=0.48\textwidth]{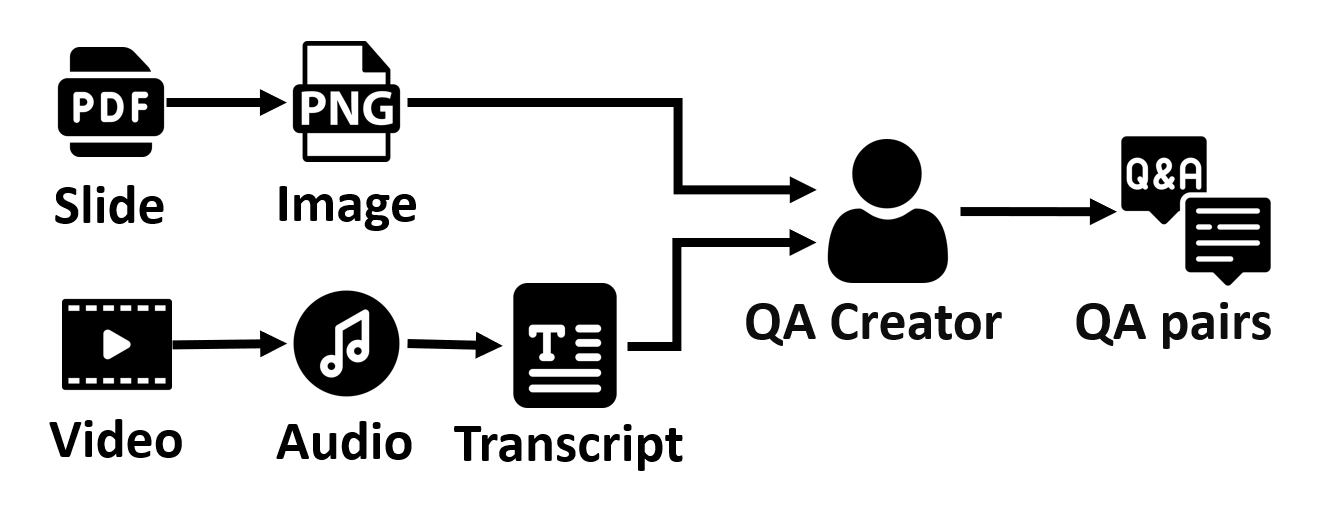} 
  \vspace{-0.6cm}
  \caption{The workflow of the data collection process.}
  \label{fig:workflow}
\end{figure}

\section{Dataset Genesis and Statistical Overview}
\subsection{Data Collection}

The data collection process involves three key stages: the collection of slide images, transcriptions, and the formulation of questions and answers. 
Fig.~\ref{fig:workflow} shows the workflow of these stages, which start with the conversion of slides into images, formatting videos to audio, then getting the transcripts, and ending with the generation of question-answer pairs. We carefully examined each data collection stage to provide clarity on the methods and their significance in the context of machine learning analysis\footnote{The slides are the property of the teacher and the university. For the privacy issue, it cannot be tested with closed-source large language models, like ChatGPT.}.

\begin{table}[h]
    \centering
        \caption{List of machine learning course topics by week.}
    \label{tab:list_topics}
    \begin{tabular}{cc}
        \hline
     Week & Topic \\
    \hline
1 & Introduction to Machine Learning \\
2 & Basic Math Recap \& Data Preprocessing \\
3 & Classification \& Regression \\
4 & Logistic Regression Model \& Least Squares \\
5 & Principle Components Analysis \& Factor Analysis \\
6 & Matrix Factorization \\
7 & Clustering \\
8 & Gaussian Mixture Models \\
9 & K-Nearest Neighbor \\
10 & Decision Trees \\
11 & Support Vector Machines \\
14 & Vision Transformer \\
15 & Ensemble Learning \\
16 & Conclusion \\
    \hline
    \end{tabular}
\end{table}


\noindent \textbf{Image Collection.} 
We carefully collected and converted slides from the PDF presentations of a 14-week machine learning course into 885 PNG images. The PyMuPDF module~\cite{PyMuPDF} was used to transform the educational materials into a visual format. The images cover a wide range of topics, from basic machine learning concepts to advanced topics such as vision transformers and ensemble models. We present a detailed overview of the curriculum for the machine learning course, encapsulated in Tab.~\ref{tab:list_topics}.


\noindent \textbf{Transcript Collection.} 
The goal of transcript collection is to transcribe the verbal content from the recorded lecture videos of the machine learning course. The transcription tools we employed are designed to process audio files. Therefore, our initial step was to extract audio tracks from video recordings of the lectures. Once the audio was successfully extracted, we employed two main types of transcription tools: Python-based models, such as the Silero model~\cite{silero2023, silero_github}, Wav2Vec Base model~\cite{DBLP:journals/corr/abs-2006-11477}, Wav2Vec2 large-lv60 model~\cite{DBLP:journals/corr/abs-2006-11477}, and the Google Speech-to-Text API~\cite{googlesttapi}, as well as professional online platforms such as Cockatoo~\cite{cockatoo}, Deepgram~\cite{deepgram2023}, Trint~\cite{trint2023}, Parrot~\cite{parrot2023}, Veed~\cite{veed2023}, and Speechtext~\cite{speechtext2023}.

After rigorous testing of these ten tools, we found that professional platforms vastly outperformed Python-based models. The output from Wav2Vec and similar Python tools was fundamentally flawed, making the creation of coherent sentences unattainable. While professional tools mostly produced accurate transcripts, some errors occurred. These mistakes did not affect the grammar but led to factually wrong sentences, which could change the text's intended meaning. After detailed analysis and review, Deepgram~\cite{deepgram2023} emerges as the superior option with distinguished and exceptional accuracy.


The initial poor quality of the recordings presented a significant obstacle due to background noise and distance from the speaker to the microphone, which often made the audio difficult to understand. Therefore, despite time constraints, careful manual review and adjusting transcript contents to its original meanings was essential to ensure the training quality. For instance, reviewing a two-hour video could take five times that amount of time or more, underscoring the labor-intensive nature of this phase. 

\begin{table}[h]
    \centering
        \caption{QA Pair JSON Schema.}  
    \label{table:qa_schema}  
    \begin{tabular}{p{1.5cm}p{5cm}}  
        \hline
        \textbf{Field} & \textbf{Description} \\  
        \hline
Instruction & The question or instruction for the QA pair. \\
Context & Contextual information, often including slide or transcript content. \\
Response & The corresponding answer or response to the question. \\
Category & The category of the QA pair, e.g., 'closed\_qa', 'information\_extraction'. \\
Week & The week of the ML course to which the content belongs. \\
Page & The page number of the slide. \\
        \hline
    \end{tabular}
\end{table}






\noindent \textbf{Creating Questions and Answers.}
The final stage is to generate questions and answers for each slide, making sure to include contextual clues that guide the answers and metadata such as the week of the lecture and the slide number. We create about ten question-answer pairs per slide, focusing on the core question, ``Can you explain this slide?" and other types of questions listed in Tab.~\ref{tab:category_distribution} to create an extensive collection of summary questions. To maintain consistency and facilitate easy access, we meticulously arrange and store these question-answer pairs in a structured JSON format. Each JSON object is governed by the schema shown in Tab.~\ref{table:qa_schema}.


\subsection{Statistical Summary}
We name our dataset as MLVQE. This section provides a comprehensive statistical overview of our MLVQE dataset, offering insights into its composition and variability that are essential for the proposed method.

\textbf{Dataset Overview.} Tab.~\ref{tab:data_breakdown} shows the detailed composition of the dataset, highlighting the predominant presence of 9,416 question-answer pairs. These are accompanied by 110,407 words of transcripts that provide detailed textual documentation of the lectures. The dataset also includes 885 images, with each image directly linked to a specific lecture slide and its corresponding transcript.


\textbf{The Structure of the Dataset.}
The dataset consists primarily of lecture slides in a visual format. Recorded as images at a resolution of 960 by 540 pixels, these slides encapsulate the most important topic-related information from each lecture. They provide a visual overview and textual reference and convey the lecture's overarching narrative. The categorization of our textual dataset reflects the variety of questions or tasks it encompasses. According to the "Category Distribution" detailed in Tab.~\ref{tab:category_distribution}, the dataset has several distinct categories, including closed-ended questions: ``closed\_qa", information extraction: ``information\_extraction", open-ended questions: ``open\_qa", among others. The ``closed\_qa" category contains the largest number of entries, totaling 2,776, whereas the ``creative\_writing" category has the smallest, with only 326 entries.

\begin{table}[H]
\centering
\caption{Breakdown of the collected data.}
\label{tab:data_breakdown}
{\scriptsize
\begin{tabular}{@{}lc@{}}
\toprule
\textbf{Data Type} & \textbf{Count} \\
\midrule
Question-Answer pairs & 9,416 \\
Transcripts & 110,407 \\
Images & 885 \\
\bottomrule
\end{tabular}
}
  \vspace{-0.3cm}
\end{table}

\begin{table}[H]
\centering
\caption{Category Distribution.}
\label{tab:category_distribution}
{\scriptsize
\begin{tabular}{@{}lc@{}}
\toprule
\textbf{Category} & \textbf{Number} \\
\midrule
closed\_qa & 2,776 \\
information\_extraction & 2,122 \\
general\_qa & 1,125 \\
open\_qa & 1,083 \\
summarization & 934 \\
brainstorming & 540 \\
classification & 510 \\
creative\_writing & 326 \\
\bottomrule
  \vspace{-0.6cm}
\end{tabular}
}
\end{table}

\begin{table}[H]
  \centering
  
  \caption{Weekly Breakdown of QAs, Transcripts/Images, and Corresponding Transcript Word Numbers.}
  \label{tab:combined_distribution}
  {\scriptsize 
  \begin{tabular}{@{}cccc@{}} 
  \toprule
  \textbf{Week} & \textbf{QAs} & \textbf{Transcripts/Images} & \textbf{Word Count} \\ 
  \midrule
  1 & 1,090 & 92 & 9,775 \\
  2 & 765 & 68 & 9,005 \\
  3 & 1,210 & 117 & 12,228 \\
  4 & 635 & 61 & 10,289 \\
  5 & 806 & 81 & 11,147 \\
  6 & 525 & 53 & 5,051 \\
  7 & 780 & 76 & 7,590 \\
  8 & 560 & 54 & 6,195 \\
  9 & 390 & 38 & 6,386 \\
  10 & 800 & 78 & 7,785 \\
  11 & 440 & 41 & 7,804 \\
  14 & 680 & 58 & 8,047 \\
  15 & 600 & 56 & 8,193 \\
  16 & 135 & 12 & 912 \\
  \midrule
  \textbf{Total} & 9,416 & 885 & 110,407 \\ 
  \bottomrule
  \end{tabular}
  }
  \vspace{-0.3cm}
\end{table}


\begin{table}[H]
  \centering
  \caption{Statistical Summary of Word Numbers.}
  \label{tab:stat_summary}
  {\scriptsize
  \begin{tabular}{@{}lcc@{}} 
  \toprule
  \textbf{Category} & \textbf{Mean} & \textbf{Maximum} \\
  \midrule
  Transcripts & 130.07 & 1,077 \\
  Questions & 7.60 & 22 \\
  Answers & 15.85 & 98 \\
  \bottomrule
  \end{tabular}
  }
  \vspace{-0.3cm}
\end{table}

\textbf{Distribution Data per Week.} Tab.~\ref{tab:combined_distribution} provides the weekly distribution of our dataset. The fluctuation in question-answer pairs, ranging from 1,210 in the third week to a minimum of 135 in the sixteenth week, reflects the dynamic nature of the course curriculum, with a total of 9,416 pairs. Likewise, the steady count of weekly transcripts and images, which sum up to 885, highlights the equilibrium in the presentation of visual and textual materials in the course content. Weeks 12 and 13 are holiday breaks without any courses. Hence, it is a 14-week course. 

\begin{figure}[h!]
  \centering
  \includegraphics[width=0.4\textwidth]{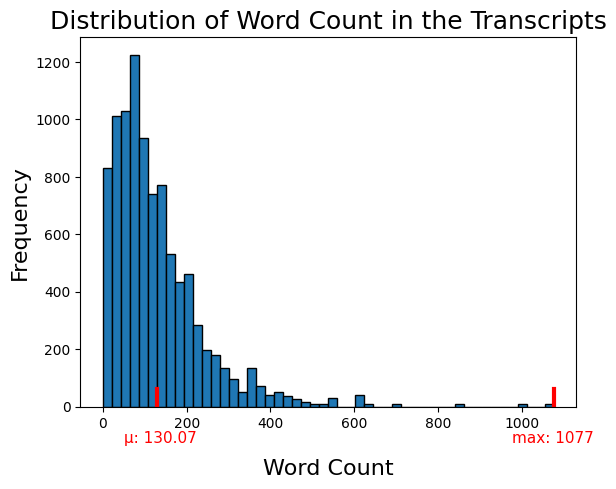} 
  \vspace{-0.3cm}
  \caption{Transcript Distribution}
  \label{fig:transcript_distribution}
\end{figure}

\begin{figure}[htbp]
\centering
  \centering
  \includegraphics[width=.8\linewidth]{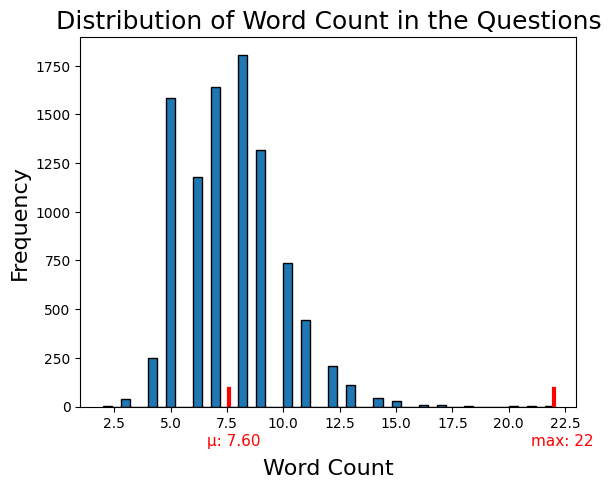}
  \vspace{-0.3cm}
  \caption{Question Distribution}
  \label{fig:question_distribution}
\end{figure}

\begin{figure}
  \centering
  \includegraphics[width=.8\linewidth]{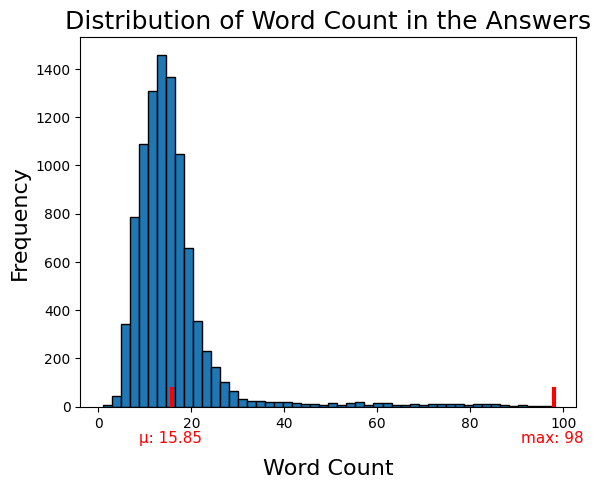}
  \caption{Answer Distribution.}
  \label{fig:answer_distribution}
\end{figure}

\textbf{Distribution of Transcript Words and  Question and Answers (QA) Pairs.}
Fig.~\ref{fig:transcript_distribution} illustrates the distribution of word counts in transcripts, highlighting the mean and maximum string lengths. Similarly, Fig.~\ref{fig:question_distribution} and Fig.~\ref{fig:answer_distribution} present the distributions of word counts in questions and answers, respectively, elucidating the typical and extreme lengths of the text being analyzed. These visual representations help underscore the variability and scope of the dataset's content, from concise questions to detailed transcripts.

%


Tab.~\ref{tab:stat_summary} offers a detailed statistical analysis of the word counts for questions, answers, and transcripts. It reveals that, on average, questions contain 7.6 words, answers comprise approximately 15.85 words, and transcripts average 130.07 words. The data also shows the range of content counts within the dataset, with maximum word counts reaching  22 for questions, 98 for answers, and 1,077 for transcripts.


\textbf{Comparative Lexical Analysis of MLVQA Datasets.}
The creation of the MLVQE dataset took \textbf{five} students for around \textbf{six months}, which is characterized by an average question length of 7.60 words and an unprecedented average answer length of 15.85 words, showcasing an unprecedented level of lexical richness and complexity. Intriguingly, for inquiries such as 'Can you explain this slide?', the average answer length escalates to 37.10 words, highlighting the dataset's capacity for detailed and comprehensive explanations. This stands in stark contrast to the Visual Genome dataset~\cite{krishna2017visual}, where the average question and answer lengths are merely 5.70 and 1.80 words, respectively. In comparison, datasets like CLEVR~\cite{johnson2017clevr} and VQAv2~\cite{goyal2017making} exhibit significantly longer questions, averaging 18.38 and 6.12 words, with answers typically spanning one to three words, thereby accentuating the pronounced distinctions. GQA~\cite{hudson2019gqa}, with its average question length of 10 words, further illustrates the variability in lexical brevity across datasets. However, the significantly verbose nature of our dataset, especially in terms of answer length, not only broadens the spectrum of language usage but also introduces a complex layer to the training of VQA models, necessitating advanced language processing capabilities. This depth and breadth of linguistic expression in our dataset challenge the development of robust models capable of nuanced understanding and generation, thereby significantly advancing the VQA field.

\section{Method}

\subsection{Motivation}

Most of the VQA models are open-ended and cannot answer domain-specific questions; they are mostly fine-tuned on downstream tasks, and in most cases, they do not answer well; it depends on how well they were trained, the data quality, and the structure. Our custom dataset MLVQE has two major aspects, transcripts, and slides, which help answer the question, "Can you explain the slide?". With the help of transcripts,  the models can predict answers like an instructor of the course. In addition, the slide and question-answer pairs are conversation-based data that help the model generate conversations.

We propose the SparrowVQE, a model innovatively trained in three sequential stages to adeptly integrate multimodal data, specifically slide-text pairs, addressing a significant gap in educational technology. 
This structured training enhances its capability in visual question answering and educational content synthesis, promising a transformative impact on dynamic learning environments.


\subsection{Model Architecture}
\begin{figure}[h]
  \centering
\includegraphics[width=1\linewidth, height=0.23\textwidth]{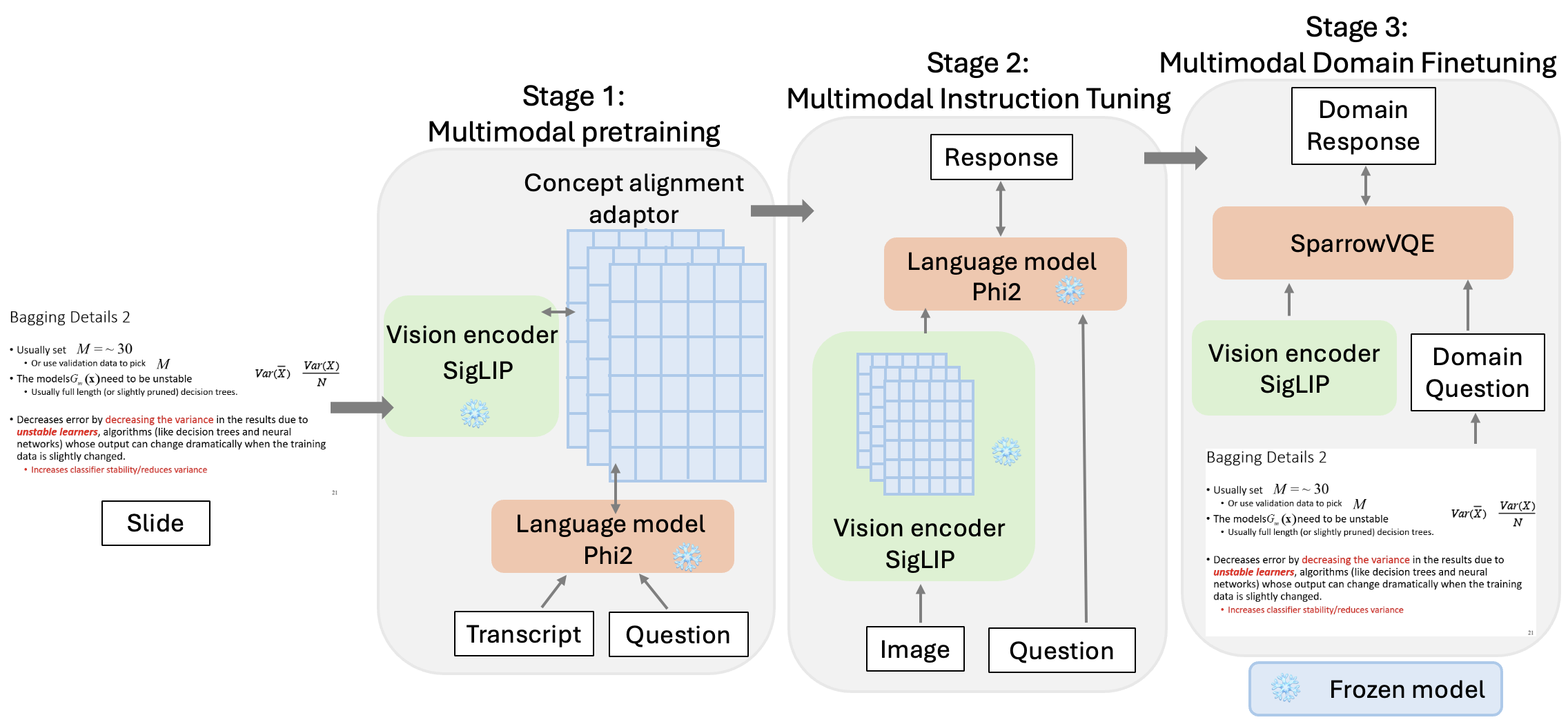}
  \caption{The three-stage training mechanism of our proposed SparrowVQE model.
  }
  \label{fig:flow}
\end{figure}

SparrowVQE connects the frozen image encoder using a trainable module of two-layer MLP~\cite{chen2020simple, chen2020improved} to a casual language model. The overall workflow is shown in Fig.~\ref{fig:flow}. The pre-trained language model is Phi2~\cite{li2023textbooks}, which is a 2.7 billion-parameter language model that demonstrates outstanding reasoning and language understanding capabilities, showcasing state-of-the-art performance among base language models with less than 13 billion parameters, and the vision encoder is Sigmoid loss for Language-ImagePre-training (SigLIP)~\cite{zhai2023sigmoid}. Unlike standard contrastive learning with softmax normalization, the sigmoid loss operates solely on image-text pairs and does not require a global view of the pairwise similarities for normalization.
This custom 3B parameter model tailored for specialized educational settings incorporates slide-text pairs from machine learning courses. 

 Model consists of multimodal pre-training (slide image and transcripts feature alignment), instruction tuning of the pre-trained model with transcripts and QA pairs, and domain fine-tuning of slide, image, and QA pairs.\subsection{Training}

\textbf{Machine Learning Concept Alignment Data.} We adopted and expanded the training pipeline and datasets from the LLaVA-v1.5~\cite{liu2023improved} framework. Our methodology worked through three training stages, each designed to enhance our model's multi-modal capabilities on our MLVQE dataset.
%
%
During the initial phase, Multimodal Pretraining, we employed a subset of slide and transcript pairs to train a concept alignment adapter. The adapter was developed to create a vector compatibility interface between a frozen pre-trained vision encoder (SigLIP) and a frozen Large Language Model (Phi-2), enabling a symbiotic processing of visual and textual information. In the subsequent phase, Multimodal Instruction Tuning involved the refinement of the adapter with more image and text pairs constituting different classes, giving the model a wider range of understanding between the images and the text pairs. This instruction tuning utilized other benchmark multimodal VQA instruction data alongside our MLVQE formatted datasets to boost the model's proficiency in complex multimodal instructions. Later, the domain adaptation was carried out from the instruction checkpoint to give the model a more robust understanding between the slide and QA pairs; details about each stage are provided in the consecutive sections.\\

\begin{table}[H]
\vspace{-0.5cm}
\centering
\caption{List of datasets used by SparrowVQE during training.}
\label{tab:datasets_stages}
{\small
\begin{tabular}{@{}lcccc@{}}
\toprule
Dataset & Stage 1 & Stage 2 & Stage 3 \\
\midrule
LLaVA-Instruct-150K\cite{liu2023llava} & $\times$  & $\checkmark$ & $\times$ \\
Slide-Transcript pairs   & $\checkmark$ & $\checkmark$& $\times$ \\
TextVQA, VQAv2\cite{balanced_vqa_v2}                 & $\times$ & $\checkmark$ & $\times$\\
Visual Genome Part1 \& part2\cite{krishna2016visual}   & $\times$ & $\checkmark$ & $\times$ \\
COCO, VQA, GQA \cite{balanced_vqa_v2}               & $\times$ & $\checkmark$& $\times$ \\
Slide-Question pairs    & $\times$ & $\times$ & $\checkmark$ \\
\bottomrule
\end{tabular}
}
\end{table}

\textbf{Stage 1: Multimodal Pretraining.} During this initial phase, we leverage the subset of slide-transcript pairs from MLVQE dataset to pre-train the Machine learning concept alignment adapter. This adapter facilitates a trainable weight matrix between the visual and textual representations by minimizing the feature alignment loss. We minimized the Mean Squared Error (MSE) loss in this stage between the projected visual and textual feature vectors to enhance the model's comprehension of the domain-specific nuances in the machine learning lectures as in Eq.~\eqref{eq:mse}:
\begin{equation}\label{eq:mse}
\mathcal{L}_{\text{MSE}} = \frac{1}{N} \sum_{i=1}^{N} (f(v_i) - g(t_i))^2,
\end{equation}
where $f(\cdot)$ and $g(\cdot)$ are the transformation functions for visual and textual features, respectively, $v_i$ and $t_i$ are the visual and textual features of the $i^{th}$ sample and $N$ is the total number of samples.

\textbf{Stage 2: Multimodal Instruction Tuning.}
In the second stage, the model undergoes further refinement using a comprehensive set of multimodal instruction data, including the LLaVA-Instruct-150K~\cite{liu2023llava}, slide transcript pairs, and additional image-text pairs from diverse sources. This stage employs a Cross-Entropy Loss function to optimize the model's performance on a wide range of VQA tasks, effectively enhancing its understanding of complex instructions and the ability to generate coherent responses, illustrated in Eq.~\eqref{eq:ce}:
\begin{equation}\label{eq:ce}
\mathcal{L}_{\text{Stage2}} = -\sum_{o=1}^{O} \sum_{c=1}^{M} y_{o,c} \log(p_{o,c}),
\end{equation}
where $O$ is the total number of observations, $M$ is the number of classes, $y_{o,c}$ is a binary indicator of whether class $c$ is the correct classification for observation $o$, and $p_{o,c}$ is the predicted probability of observation $o$ being of class $c$.
%


\textbf{Stage 3: Multimodal Domain Finetuning using PEFT and LoRA}

In the final stage, SparrowVQE leverages parameter-efficient fine-tuning (PEFT) and low-rank adaptation (LoRA) techniques to refine its predictive capabilities for visual question explanation. PEFT aims to reduce the computational and memory requirements of fine-tuning large language models by introducing a small set of trainable parameters. The original model parameters remain frozen, and fine-tuning is performed on the added parameters. By employing PEFT and LoRA techniques in stage 3, SparrowVQE can leverage the benefits of efficient fine-tuning while maintaining or enhancing the model's performance on the VQE task. 


The model undergoes a specialized fine-tuning process employing Parameter Efficient Fine-Tuning (PEFT) and Low-Rank Adaptation (LoRA). The original model parameters, denoted by $\theta$, are enhanced with an additive update $\Delta_{\phi}(\theta)$ derived from the PEFT parameters ${\phi}$ resulting in the updated model parameters $\theta'$. This PEFT function computes updates informed by both the current model state $\theta$ and the newly introduced PEFT parameters ${\phi}$.

LoRA, a variant of PEFT, optimizes the fine-tuning by introducing low-rank matrices B and A, which are smaller in size yet capture the essence of the update needed for the weights W in the model. These matrices are multiplied to produce the update ${BA}$, which is then added to the original weight matrix 
${W}$
to produce the updated weight matrix 
${W'}$. This low-rank update not only makes the fine-tuning process more parameter-efficient but also helps in retaining the original model's structure while allowing for significant adaptations. Using LoRA is particularly effective as it balances the fine-tuning's specificity and efficiency, making it a powerful tool for domain-specific adaptations.

Let $\theta$ be the original model parameters and $\phi$ be the added PEFT parameters. The updated model parameters $\theta'$ can be computed as:
\begin{equation}\label{eq:theta}
\theta' = \theta + \Delta_{\phi}(\theta),
\end{equation}
where $\Delta_{\phi}(\theta)$ is a PEFT function that computes the updates based on $\theta$ and $\phi$.
LoRA is a specific implementation of PEFT that introduces low-rank matrices to adapt the original model. It decomposes the weight update into two low-rank matrices, resulting in a more efficient and effective fine-tuning process.

Let $W$ be a weight matrix in the original model. The updated weight matrix $W'$ in LoRA can be expressed as:

\begin{equation}
W' = W + BA,
\end{equation}
where $B$ and $A$ are low-rank matrices, and their product $BA$ represents the weight update.

Upon completing stage 3 training, we must integrate the newly trained parameters with the foundational model established in stage 2. This integration will enhance the model's capability to discern the nuanced differences between slides and question-answer pairs, significantly improving its performance. As a result, the model will achieve precise predictions for the MLVQE dataset, demonstrating a comprehensive understanding of the domain.

\begin{figure*}[h!]
  \vspace{-3mm}
  \centering
  \includegraphics[width=0.85\textwidth,height=0.4\textheight,keepaspectratio]{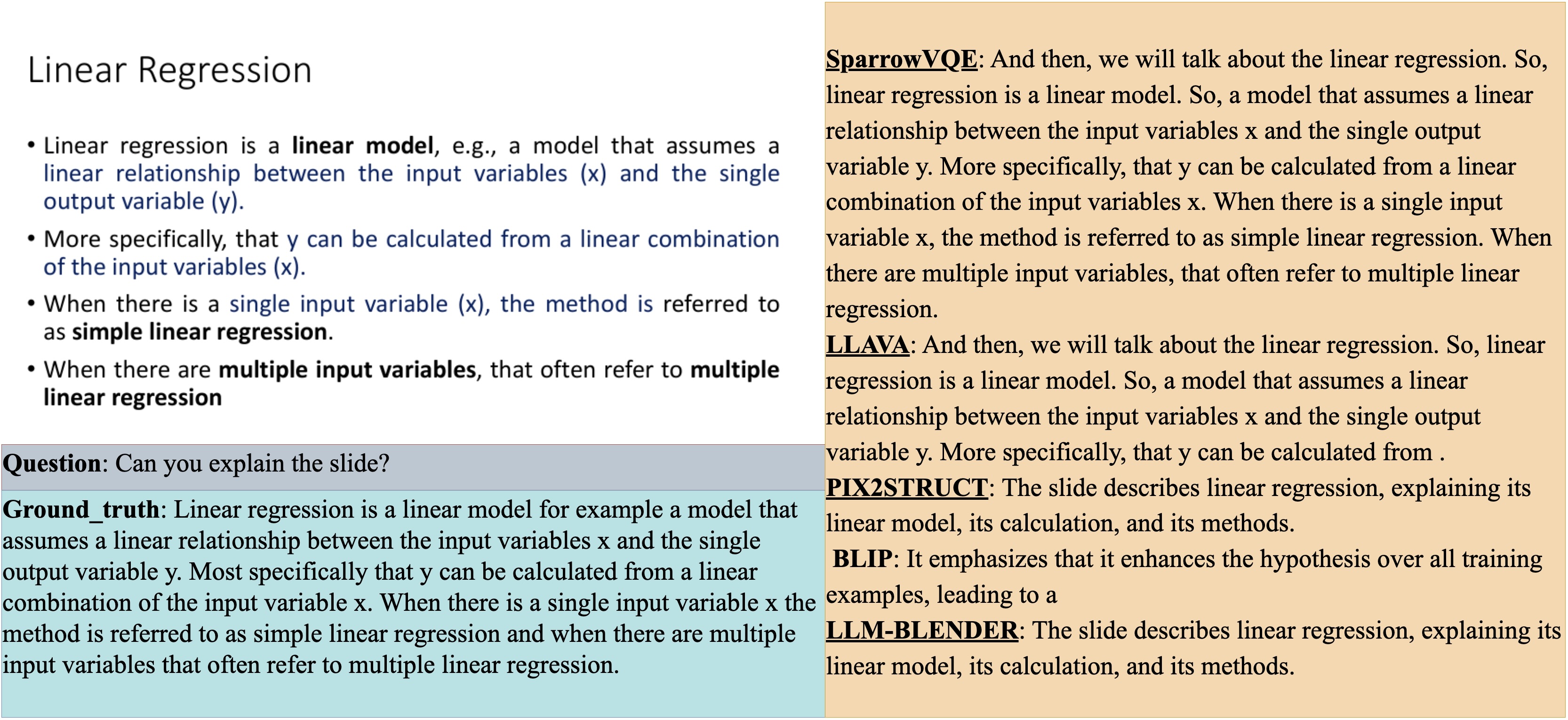}
  \caption{Results comparisons on the question "Can you explain the slide?" on Linear Regression concept from our MLVQE dataset.}
  \label{fig:comparison_random_qna}
  \vspace{-0.3cm}
\end{figure*}

\begin{algorithm}[H]
\caption{Three-Stage Training Process for SparrowVQE}
\label{alg:SparrowVQE}
\begin{algorithmic}[1]
\STATE {\bfseries Stage 1: Multimodal Pretraining}
\STATE {\bfseries Input:} Slide-transcript pairs from MLVQE dataset
\STATE {\bfseries Output:} Trained concept alignment adapter $F(\theta)$
\FOR{$s_{i,j} = S_{i,j} - 1, ..., 1$}
    \STATE Extract visual features $v = f(x)$
    \STATE Extract textual features $t = g(y)$
    \STATE Align features using (Eq.~\eqref{eq:mse})
    \STATE Update $\theta$ using AdamW: $\theta \leftarrow \text{AdamW}(\theta, \nabla_\theta \mathcal{L}, \text{lr}, \text{weight\_decay})$
\ENDFOR
\STATE {\bfseries Stage 2: Multimodal Instruction Tuning}
\STATE {\bfseries Input:} Slide-Question pairs from MLVQE and LLaVA-Instruct dataset
\STATE {\bfseries Output:} Instruction Tuned SparrowVQE model $G(\theta)$
\FOR{$q_{i,j} = Q_{i,j} - 1, ..., 1$}
    \STATE Generate predictions $p_{i,c} = G(q_i, q_j, \theta)$
    \STATE Compute Cross-Entropy Loss for batch $i$ using (Eq.~\eqref{eq:ce})
\ENDFOR
\STATE {\bfseries Stage 3: Multimodal Domain Finetuning}
\STATE {\bfseries Input:} Slide-Conversation pairs from MLVQE dataset
\STATE {\bfseries Output:} Domain Finetuned SparrowVQE model $S(\theta)$
\FOR{each slide-conversation pair $(x_{\text{slide}}, Y_{\text{conv}})$}
    \FOR{each question $y_q \in Y_{\text{conv}}$}
        \STATE Encode slide $x_{\text{slide}}$ and question $y_q$ using $S(\theta)$
        \STATE Apply LoRA for adjusted parameters $\theta'$ and weights $W'$
        \STATE Generate prediction $p_{y_q} = S(x_{\text{slide}}, y_q, \theta')$
        \STATE Compute loss using updated parameters and weights
        \STATE Accumulate gradients for $\theta'$ and $W'$
    \ENDFOR
    \STATE Update $\theta'$ and $W'$: $\theta' \leftarrow \text{AdamW}(\theta', \nabla_{\theta'} \mathcal{L}, \text{lr}, \text{weight\_decay})$, $W' \leftarrow \text{UpdateLoRA}(W', \nabla_{W'} \mathcal{L})$
\ENDFOR
\end{algorithmic}
\end{algorithm}

This comprehensive training methodology, distributed across three strategic stages, significantly elevates SparrowVQE's capability to process and interpret multimodal data, ensuring its adept at handling the complexities of machine learning education content. The training strategy, as shown in Tab.~\ref{tab:datasets_stages}, goes beyond simple fine-tuning. Typically, models are trained to answer questions based on a dataset, but our goal was to emulate the teaching style of a professor. Our three-stage method successfully enhanced the model's performance on our MLVQE dataset. 

\begin{figure*}[h!]
  \centering
\includegraphics[width=0.95\textwidth,height=0.5\textheight,keepaspectratio]{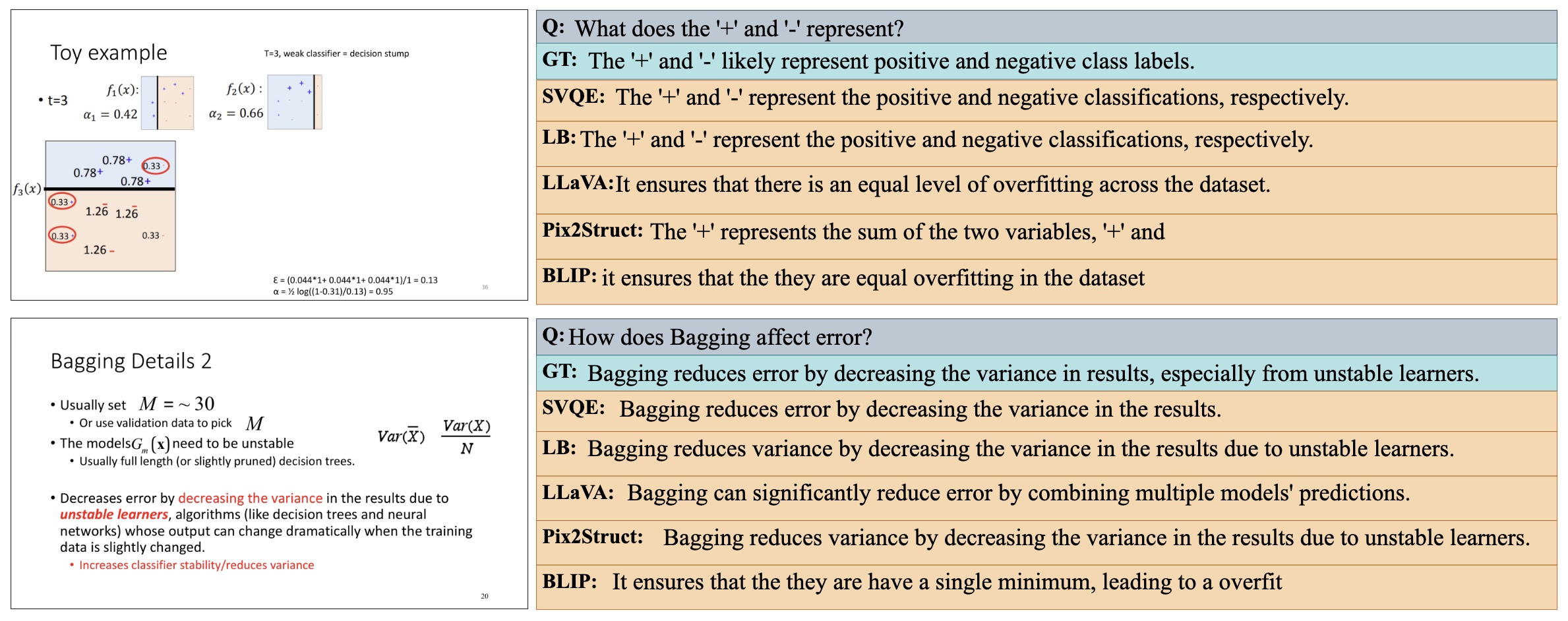}
  \caption{Comparison of different models. GT refers to ground truth, SVQE refers to SparrowVQE, LB refers to LLM-Blender. It is evident that SVQE is performing better than the other models we have experimented.}
  \label{fig:example_1}
  \vspace{-0.3cm}
\end{figure*}

\section{Experiments}
\subsection{Evaluation metrics}

We evaluate the following seven metrics on our developed MLVQE dataset. ROUGE\cite{ganesan2018rouge} assesses the similarity between predicted and ground truth answers in VQA tasks, focusing on unigrams, bigrams, and longest subsequences. BLEU~\cite{10.3115/1073083.1073135} measures the precision of n-gram overlaps in predicted answers. METEOR~\cite{10.5555/1626355.1626389} evaluates answers for linguistic variations like synonyms and paraphrases, aligning with human judgment. CIDEr~\cite{vedantam2015cider} uses TF-IDF~\cite{ref1} to gauge the relevance and uniqueness of VQA predictions compared to the ground truth. COSINE~\cite{7577578} score calculates the cosine of the angle between the vectors of the two answers in a high-dimensional space. The higher these metrics, the better the model is.


\begin{table}[H]
\vspace{-0.3cm}
\centering
\caption{Parameters of our SparrowVQE.}
\label{tab:training_hyperparameters}
{\small
\begin{tabular}{@{}lcccc@{}}
\toprule
Hyperparameter & Stage 1 & Stage 2 & Stage 3 & LoRA \\ 
\midrule
Batch size & 256 & 128 & 128 & 128\\
Learning rate (lr) & 1e-3 & 2e-5 & 2e-4 & 2e-5\\
Epoch & 2 & 2 & 10 & - \\
DeepSpeed & 2 & 3 & 3 & 3\\
\bottomrule
\end{tabular}
}
\end{table}

\textbf{Implementation details}. We train our SparrowVQE using 8 A100 80GB GPU with a three-stage training scheme. Across all three stages of training, the computation time is  5, 9, and 3 hours, respectively. We used the following parameters: a learning rate (LR) scheduler employing cosine decay, an LR warmup ratio set to 0.03, an AdamW optimizer, and a weight decay of 0.0. Tab.~\ref{tab:training_hyperparameters} shows other parameters. We utilized the first 12 weeks as the training set (8,681 QA pairs) and the remaining two weeks (735 QA pairs) as the test dataset.


\subsection{Results}
Fig.~\ref{fig:comparison_random_qna} presents the comparison of predicted results of the remarkable questions, ``Can you explain this slide?". The results show the superiority of our model in the capacity of slide information explanation. 
Fig.~\ref{fig:example_1} shows the comparison of predicted results from different models based on two proposed questions from respective lecture slides. Among the predicted answers of all models, our SparrowVQE model produced the answers closest to the ground truth answers, showing the outstanding performance of our model compared to others. 
The results on the training set and the test set in Tab.~\ref{tab:train_performance} and Tab.~\ref{tab:test_performance} show that our SparrowVQE model outperforms all other models, including BLIP~\cite{li2023blip2}, Pix2Struct~\cite{lee2023pix2struct}, LLaVA~\cite{liu2023llava}, and LLM-Blender~\cite{llm-blender-2023}, on all metrics of the answer-predicting task and reference coherence. In Tab.~\ref{tab:model_comparison}, we compare the performance of our model with LLaVA-1.5~\cite{liu2023improved}, LLaVA-Phi~\cite{li2023textbooks}, MobileVLM~\cite{chu2023mobilevlm},  MC-LLaVA-3b~\cite{lin2024rethinking}, and TinyGPTV~\cite{yuan2023tinygpt} on multiple datasets and compare their performance. It turns out that the SparrowVQE model not only stands out from other models on the question-answering performance but also presents better robustness on different benchmark datasets.

\begin{table}[h!]
  \centering
  \caption{Comparative Analysis on the TRAIN Subset of the MLVQE Dataset.}
  \label{tab:train_performance}
  \resizebox{0.5\textwidth}{!}{
  \begin{tabular}{@{}lcccccccc@{}}
    \toprule
    Models & Size & Rouge-1 & Rouge-2 & Rouge-L & COSINE & BLEU & CIDEr & METEOR \\
    \midrule
    BLIP\cite{li2023blip2}    & 224M & 8 & 0.7 & 7.5 & 0.089 & 0.15 & 0.17 & 0.079  \\
    Pix2Struct~\cite{lee2023pix2struct} & 1.3B & 43.3 & 26.6 & 49.19 & 0.41 & 0.43 & 0.49 & 0.408 \\
    LLaVA~\cite{liu2023llava}  & 7B & 41.09 & 24.2 & 38.7 & 0.406 & 0.41 & 0.57 & 0.452 \\
    LLM-Blender\cite{llm-blender-2023} & 124M & 59.8 & 44 & 57.4 & 0.570 & 0.601 & 0.62 & 0.604 \\
    \midrule
    SparrowVQE  & 3B & \textbf{65.2} & \textbf{51.84} & \textbf{58.54} & \textbf{0.60} & \textbf{0.612}& 
    \textbf{0.634} & 
    \textbf{0.610} \\
    \bottomrule
  \end{tabular}
  }
\end{table}

\begin{table}[h!]
  \centering
  \caption{Comparative Analysis on the TEST Subset of the MLVQE Dataset.}
  \label{tab:test_performance}
  \resizebox{0.5\textwidth}{!}{
  \begin{tabular}{@{}lcccccccc@{}}
    \toprule
    Models & Size & Rouge-1 & Rouge-2 & Rouge-L & COSINE & BLEU & CIDEr & METEOR \\
    \midrule
    BLIP\cite{li2023blip2}   & 224M & 8.4 & 0.7 & 7.19 & 0.077 & 0.15 & 0.17 & 0.078 \\
    Pix2Struct~\cite{lee2023pix2struct} & 1.3B & 38 & 20.1 & 35.5 & 0.365 & 0.4 & 0.47 & 0.379 \\
    LLaVA\cite{liu2023llava}  & 7B & 35.4 & 18.4 & 33.0 & 0.34 & 0.37 & 0.53 & 0.42 \\
    LLM-Blender\cite{llm-blender-2023} & 124M & 51.5 & 34.8 & 49 & 0.489 & 0.54 & 0.573 & 0.573 \\
    \midrule
    SparrowVQE & 3B & \textbf{68.13} & \textbf{51.54} & \textbf{63.92} & \textbf{0.61} & \textbf{0.7} & \textbf{0.67} & \textbf{0.652} \\
    \bottomrule
  \end{tabular}
  }
\end{table}

\begin{table}[h!]
  \centering
  \caption{Comprehensive cross-dataset performance of different models and datasets.}
  \label{tab:model_comparison}
  \resizebox{0.5\textwidth}{!}{
  \begin{tabular}{@{}lccccccc@{}}
    \toprule
    Models & Size & VQAv2~\cite{goyal2017making} & MLVQE & VizWiz & SQA & TextVQA & MMB  \\
    \midrule
    LLaVA-1.5~\cite{liu2023improved} & 7B & 78.5 & 40.2 & 50 & 66.8 & 58.2 & 64.3 \\
    LLaVA-Phi~\cite{li2023textbooks} & 3B & 71.4 & 14.6 & 35.9 & 68.4 & 48.6 & 59.8  \\
    MobileVLM~\cite{chu2023mobilevlm} & 3B & - & 10.7 & - & 61 & 47.5 & 59.6 \\
    MC-LLaVA-3b~\cite{lin2024rethinking} & 3B & 76.72 & 22.1 & 24.88 & - & 38.59  & - \\
    TinyGPTV~\cite{yuan2023tinygpt} & 3B & - & 18.4 & 24.8 & - & - & - \\
    \midrule
    SparrowVQE & 3B & \textbf{79.99} & \textbf{65.2} & \textbf{50.32} & \textbf{68.42} & \textbf{61.25} & \textbf{68.38}\\
    \bottomrule
  \end{tabular}
  \vspace{-0.3cm}
  }
\end{table}

  \section{Discussion}

\begin{table}[h!]
\centering
\caption{Ablation study of different stages on MLVQE dataset.}
\label{tab:ablation_performance}
\resizebox{0.5\textwidth}{!}{ 
  \begin{tabular}{@{}lccccccc@{}}
    \toprule
    Stages & Rouge-1 & Rouge-2 & Rouge-L & COSINE & BLEU & CIDEr & METEOR \\
    \midrule
    Stage 2  & 15 & 14.7 & 19 & 0.14 & 0.21 & 0.16 & 0.22\\
    Stage 3  & 26.6 & 25 & 28 & 0.35 & 0.30 & 0.40 & 0.39\\
    1$+$2  & 32.6 & 40.4 & 52 & 0.39 & 0.55 & 0.58 & 0.49\\
    1$+$3  & 12.4 & 10 & 25 & 0.20 & 0.25 & 0.26 & 0.54 \\
    2$+$3  & 42.4 & 30 & 45 & 0.40 & 0.45 & 0.56 & 0.54 \\
    \midrule
    1$+$2$+$3  & \textbf{68.13} & \textbf{51.54} & \textbf{63.92} & \textbf{0.61} & \textbf{0.7} & \textbf{0.67} & \textbf{0.652}\\
    \bottomrule
  \end{tabular}
}
\vspace{-0.3cm}
\end{table}

\begin{table}[h!]
  \centering
  \caption{Generation results comparison of deep learning Dataset.}
  \label{tab:test_performance2}
    \vspace{-0.3cm}
  \resizebox{0.5\textwidth}{!}{
  \begin{tabular}{@{}lcccccccc@{}}
    \toprule
    Models & Size & Rouge-1 & Rouge-2 & Rouge-L & COSINE & BLEU & CIDEr & METEOR \\
    \midrule
    BLIP\cite{li2023blip2}   & 224M & 1.3 & 1.2 & 3.6 & 0.092 & 0.16 & 0.002 & 0.028 \\
    Pix2Struct~\cite{lee2023pix2struct} & 1.3B  & 8.1 & 3.1 & 7.3 & 0.128 & 0.29  & 0.007 & 0.348\\
    LLaVA\cite{liu2023llava}  & 7B & 37.1 & 15.0 & 35.4 & 0.346 & 0.32 & 0.325 & 0.359 \\
    LLM-Blender\cite{llm-blender-2023} & 124M & 28.7 & 9.1 & 20.9 & 0.102 & 0.18 & 0.178 & 0.296\\
    \midrule
    SparrowVQE & 3B & \textbf{38.5} & \textbf{17.1} & \textbf{36.27} & \textbf{0.478} & \textbf{0.43} & \textbf{0.525} & \textbf{0.463} \\
    \bottomrule
  \end{tabular}
  }
\vspace{-0.3cm}
\end{table}

\textbf{Ablation Study.} Our SparrowVQE model outperforms state-of-the-art models in VQA tasks because of our unique multi-stage training. We demonstrate the contribution of each of our training stages toward the overall efficacy and accuracy of our model with the ablation study on our proposed MLVQE dataset. Our model has three sequential training stages: multimodal pre-training for aligning slide image features with transcripts (Stage 1), instruction tuning with transcript and QA pairs to adapt the pre-trained model to our specific tasks (Stage 2), and domain-specific fine-tuning with slide image and QA pairs to further refine the model's performance in our target domain (Stage 3). Stage 1 is necessary to train our SparrowVQE model; hence, we exclude the performance of Stage 1 alone. As shown in Tab.~\ref{tab:ablation_performance} (1: stage 1, 2: stage 2 and 3: stage 3), we can find that with the increase of stages, the performance of our SparrowVQE improves. In addition, Stage 3 gains more performance improvement than Stage 2. These results reveal that all three proposed stages are useful in improving the model's performance on the MLVQE dataset. 


\textbf{Generation to deep learning course.}
To demonstrate the generation ability of our proposed SparrowVQE model, we also tested our model on an unseen deep learning course. 
Tab.~\ref{tab:combined_distribution} lists the weekly breakdown of QAs, transcripts/images, and corresponding transcript word counts. In this new deep learning course, we have a total of 11,294 QA pairs and a total of 1,177 slide images. Tab.~\ref{tab:test_performance} shows the comparison results of four baseline models with our SparrowVQE model. For all models, we only make the inference for the QA pairs without training. We can find that our SparrowVQE model has better performance in all of the seven metrics, including Rouge-1, Rouge-2, Rouge-L, COSINE, BLEU, CIDEr, and METEOR. This result reveals that our model has a higher generalization ability in the unseen deep learning course than all other models. 

SparrowVQE is finely tuned for machine learning course content, which can improve performance and contribution in potential educational settings. The model is designed to leverage the state-of-the-art linguistic capability of Phi2 for reasoning and language understanding without the computational overhead. Using the Sigmoid loss via the SigLIP vision encoder also negates the need for global normalization, leading to more efficient and scalable training processes. However, the small, 3B, size of SparrowVQE may reflect a scope of knowledge or understanding smaller than models with larger sizes, making it less capable when dealing with out-of-distribution data. Additional training is also required when applied to education settings other than machine learning, such as chemistry, finance, literacy, etc.

\textbf{Future direction.}
We plan to expand our dataset to include additional AI topics such as deep learning, natural language processing, reinforcement learning, etc. Future improvements will include adjusting the concept alignment adapter weights and applying a domain-specific fine-tuning strategy. This innovative training technique positions us at the forefront of developing models trained on slide-text pairs, offering a significant advantage in the field.

\section{Conclusion}
In this paper, we introduce a SparrowVQE model for the visual question explanation (VQE) task in educational settings. The model is designed to incorporate Phi2 and SigLIP for textual and image feature processing, with the sigmoid loss operating solely on image-text pairs. The comparison results show the superiority of SparrowVQE over other state-of-the-art models in scene-based question-answering and visual and textual understanding on our  MLVQE dataset and five other benchmark VQA datasets. In the future, we plan to adjust the concept alignment adapter and implement a domain-specific fine-tuning strategy to improve the model performance and expand its application to broader topics.

{\small
\bibliographystyle{unsrt}
\bibliography{reference}
}

\end{document}